\documentclass[journal,twoside,web]{ieeecolor}
\usepackage{jsen}
\usepackage{amsmath,amssymb,amsfonts}
\usepackage{graphicx}
\usepackage{textcomp}
\usepackage{wrapfig}

\usepackage{color}    
\usepackage{multirow}
\usepackage{amsmath}
\usepackage{amsfonts}
\usepackage{amssymb}
\usepackage{array}
\usepackage[caption=false,font=normalsize,labelfont=sf,textfont=sf]{subfig}
\usepackage{textcomp}
\usepackage{stfloats}
\usepackage{url}
\usepackage{verbatim}
\usepackage{graphicx}
\usepackage{caption}
\usepackage{bbding}
\usepackage{amssymb}
\usepackage{algorithm}
\usepackage{algpseudocode}
\usepackage{booktabs}
\DeclareUnicodeCharacter{FF1A}{\textasciitilde}
\usepackage{makecell}
\usepackage{cite}
\makeatletter
\let\NAT@parse\undefined  
\makeatother
\usepackage[colorlinks, linkcolor=blue, citecolor=blue]{hyperref}  

\usepackage{adjustbox}

\def\BibTeX{{\rm B\kern-.05em{\sc i\kern-.025em b}\kern-.08em
    T\kern-.1667em\lower.7ex\hbox{E}\kern-.125emX}}
\definecolor{abstractbg}{rgb}{0.89804,0.94510,0.83137}
\begin{document}
\title{DroneSR: Rethinking Few-shot Thermal Image Super-Resolution from Drone-based Perspective}
\author{Zhipeng Weng, Xiaopeng Liu, Ce Liu, Xingyuan Guo, Yukai Shi, and Liang Lin, \IEEEmembership{Fellow, IEEE}
\thanks{Z. Weng, X. Liu, C. Liu, and Y. Shi are with School of Information Engineering, Guangdong University of Technology, Guangzhou, 510006, China. (email:  {wzpp24@foxmail.com; xiaopeng22@foxmail.com; liuce382@foxmail.com; ykshi@gdut.edu.cn}).(Corresponding author: Yukai Shi) }
\thanks{X. Guo is with Guangdong Power Grid, Ltd., Guangzhou, 510000, China. (email: { gxypower@foxmail.com}). }
\thanks{L. Lin is with School of Computer Science, Sun Yat-sen University, Guangzhou, 510006, China. (email: { linliang@ieee.org}).}}

\IEEEtitleabstractindextext{%
\fcolorbox{abstractbg}{abstractbg}{%
\begin{minipage}{\textwidth}%
\begin{wrapfigure}[12]{r}{2in}%
\includegraphics[width=2in]{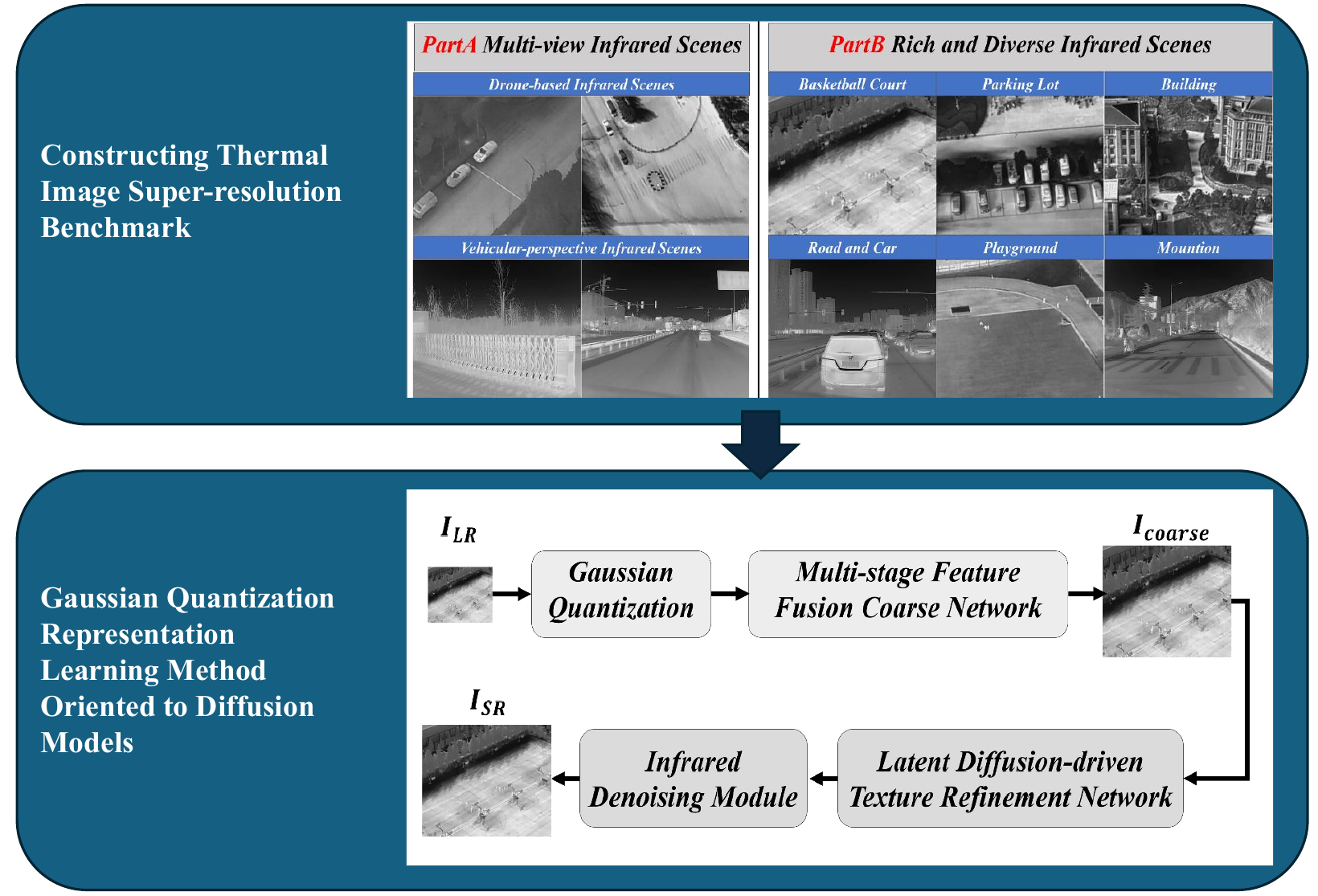}%
\end{wrapfigure}%
\begin{abstract}
Although large scale models achieve significant improvements in performance, the overfitting challenge still frequently undermines their generalization ability. In super resolution tasks on images, diffusion models as representatives of generative models typically adopt large scale architectures. However, few-shot drone-captured infrared training data frequently induces severe overfitting in large-scale architectures. To address this key challenge, our method proposes a new Gaussian quantization representation learning method oriented to diffusion models that alleviates overfitting and enhances robustness. At the same time, an effective monitoring mechanism tracks large scale architectures during training to detect signs of overfitting. By introducing Gaussian quantization representation learning, our method effectively reduces overfitting while maintaining architecture complexity. On this basis, we construct a multi source  drone-based infrared image benchmark dataset for detection and use it to emphasize overfitting issues of large scale architectures in few sample, drone-based diverse drone-based image reconstruction scenarios. To verify the efficacy of the method in mitigating overfitting, experiments are conducted on the constructed benchmark. Experimental results demonstrate that our method outperforms existing super resolution approaches and significantly mitigates overfitting of large scale architectures under complex conditions. The code and droneSR dataset will be available at: \href{https://github.com/wengzp1/GARLSR}{https://github.com/wengzp1/GARLSR}.
\end{abstract}

\end{minipage}}}

\maketitle

\section{Introduction}
\label{sec:introduction}
\IEEEPARstart{D}{DPMs}(Denoising Diffusion Probabilistic Models)~\cite{ddpm1,ddpm2,ddpm3,ddpm4} have demonstrated outstanding performance in the field of image generation in recent years. Super-resolution methods based on diffusion models generate high-resolution images through progressive sampling, allowing them to better capture real details. StableSR~\cite{stablesr} uses a pre-trained stable diffusion model as a generative prior to generate high-resolution images with strong realism from low-resolution images. DiffBIR~\cite{diffbir} combines diffusion models and a staged image processing strategy to generate restored images with rich details and high visual quality. However, large-scale models face a problem when applied to drone-based infrared super-resolution tasks: they are prone to overfitting when training on a small and complex dataset. 

\begin{figure}[!t]
\centering
\includegraphics[width=0.99\linewidth]{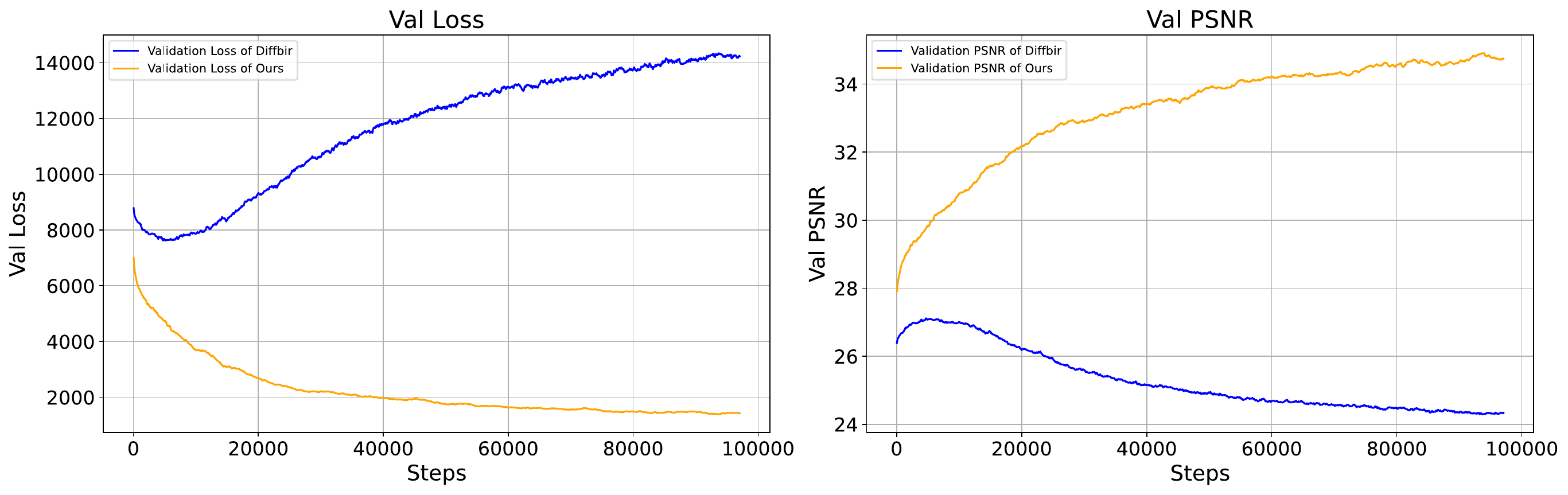}
\caption{Typical image Super-resolution methods assume that there is an ample amount of training data. However, it is expensive to capture a large amount of high-quality infrared data on drones with a target prior. Given limited training images under drone-based perspective, it can be observed that large model methods (e.g, DiffBIR~\cite{diffbir}) clearly show overfitting during the few-shot condition. Our method aim at showing a stable performance under drone-based perspective.}
\label{fig1}
\end{figure}

As shown in Fig.~\ref{fig1}, we visualize the validation loss and validation evaluation metrics Peak Signal-to-Noise Ratio (PSNR) of the diffusion model based on DiffBIR~\cite{diffbir} and our method during training, respectively. We can observe that DiffBIR~\cite{diffbir} (blue curve) begins overfitting on drone data after 10000 steps, while our method (yellow curve) maintains convergence throughout the entire training process. During the inference phase, due to overfitting in DiffBIR~\cite{diffbir}, the obtained quantitative metrics are lower than those of our method.

Infrared imaging technology is widely used in fields such as temperature detection, fault diagnosis, and remote sensing~\cite{dnn2,hung2025tinyfl_hkd}. Compared to visible light imaging, the quality of infrared images is generally lower~\cite{infrared5,infrared6}. This is due to the lower frequency of the infrared spectrum and the limited size of the focal plane array in infrared detector designs. To address this issue, researchers have introduced image super-resolution technology to reconstruct high-resolution images from low-resolution images. For example, DESRA~\cite{desra} improves the image quality and visual consistency of real-world super-resolution models by detecting and removing artifacts generated by Generative Adversarial Networks (GANs). However, in the field of infrared image super-resolution, despite many attempts using Deep Neural Network (DNN) methods~\cite{dnn2,dnn3,dnn4,dnn1}, the visual quality in practical applications still falls short. The main reasons include the following two challenges: (1) Non-blind image property: existing methods are often based on non-blind assumptions and cannot handle complex degradations and multi-source sensor patterns in practical applications. (2) Few-shot data scale: it is very difficult to collect large amounts of data for each sensor, especially for complex sensor models.

\begin{figure*}[t]
\centering
\includegraphics[width=0.95\linewidth]{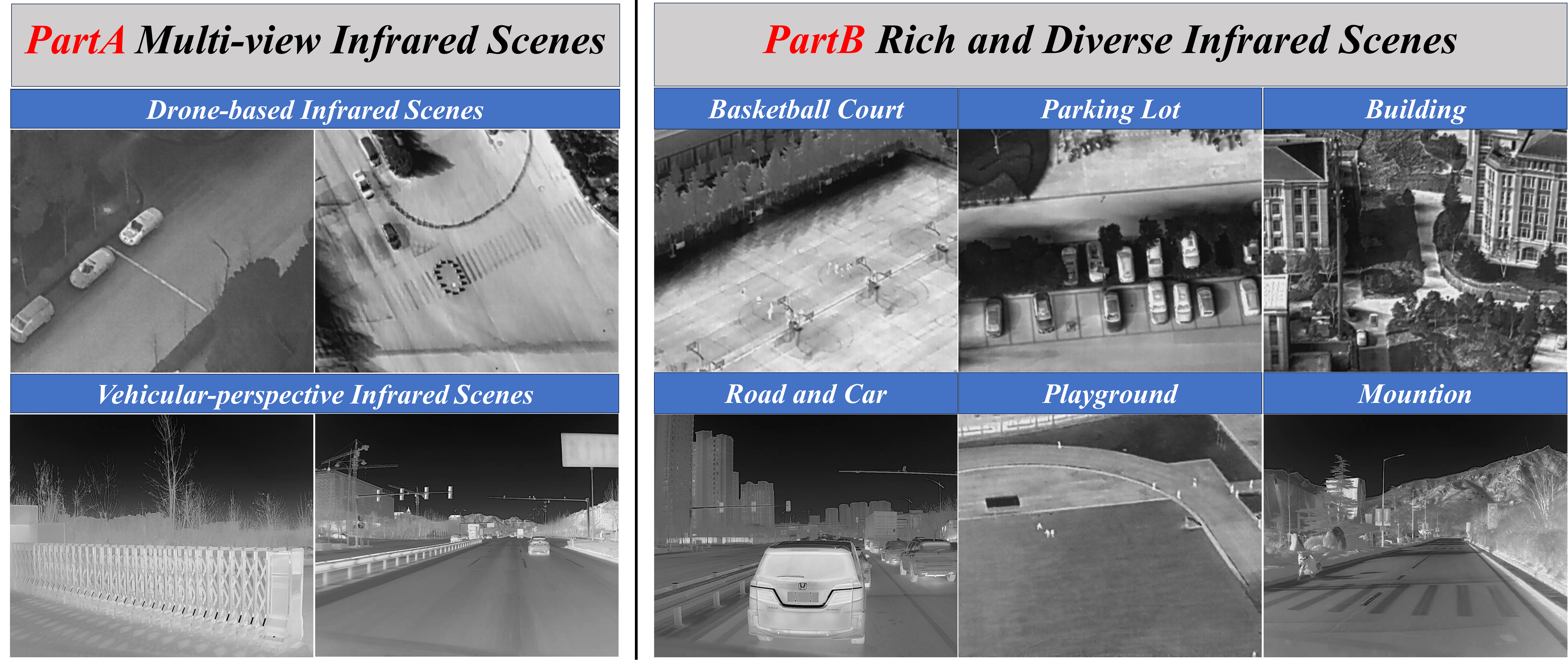}
\caption{A demonstration of thermal imaging super-resolution benchmark. We have gathered complex and diversified infrared imagery, consolidating them into an integrated infrared dataset named DroneSR, to verify current thermal image SR methods under practical scenes.}
\label{figdataset}
\end{figure*}

To address the above issues, we propose a super-resolution method based on Gaussian probability-guided quantization. It is designed to address the challenges of limited data and large-scale model parameters in infrared super-resolution tasks. Specifically, we first propose an adaptive quantization method based on Gaussian probability guidance. It performs degradation operations on the input infrared images during the data processing process. This method can reduce noise interference in low-quality infrared images and smooth the pixel distribution, while preserving the basic structural information of the infrared images. Subsequently, we input the Gaussian quantized images into the multi-stage feature fusion coarse network and the latent diffusion-driven texture refinement network to improve image quality. Then we introduce the infrared denoising module to mitigate residual noise interference. As depicted in Fig.~\ref{figdataset}, we have created a multi-source infrared dataset to effectively highlight the diversity of infrared imaging data and the limitation of data volume. Finally, we compare our method with currently advanced image super-resolution methods, and the performance metrics demonstrate the superiority of our method. In summary, the contributions of the paper are as follows:
\begin{itemize}

\item[$\bullet$]We build a new drone-captured thermal imaging super-resolution benchmark. It is designed to highlight the challenges of aerial infrared data scarcity, drone-specific scene diversity, and large-scale model optimization under drone-based infrared SR scenarios.

\item[$\bullet$]We propose a new Gaussian quantization representation learning method to address overfitting challenges in generative methods. By introducing Gaussian quantization during data processing to generate diversified low-quality infrared images, it effectively addresses the overfitting issue in large-scale models and enhances the model robustness.

\item[$\bullet$]We introduce a super-resolution restoration module based on a two-stage process. We perform an initial reconstruction of the input low-quality infrared images to generate coarse recovery images with global consistency. Subsequently, leveraging the generative potential of denoising diffusion models, we generate high-resolution and realistic images through progressive denoising and detail enhancement. 

\end{itemize}

\section{RELATED WORK}
\label{related work}
\subsection{Image Super-Resolution}

\begin{figure*}[!t]
\centering
\includegraphics[width=0.95\linewidth]{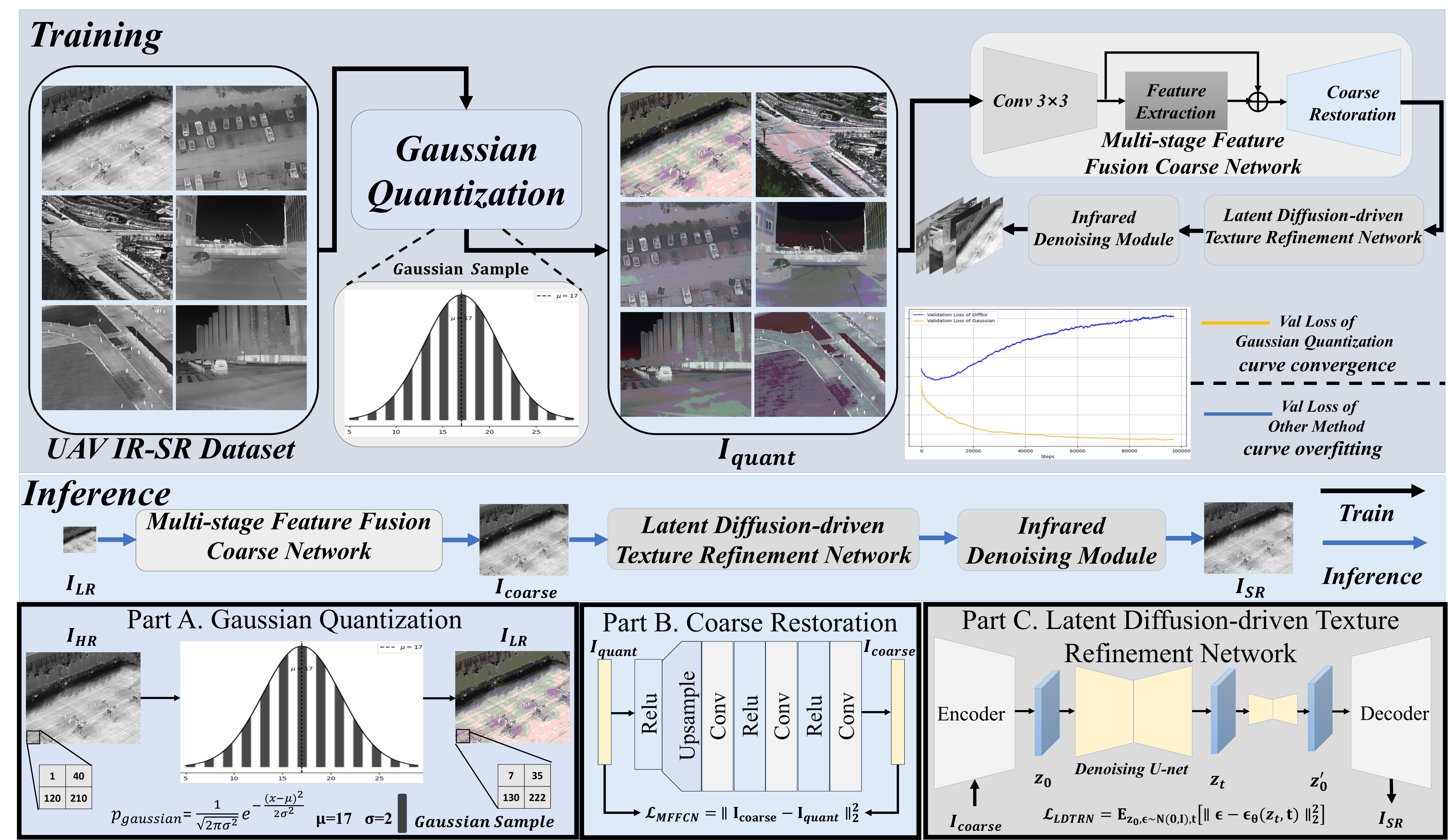}
\caption{In the training phase, our method first performs Gaussian probability guided quantization on the infrared images. Then, the quantized images $I_{quant}$ are sequentially input into the multi-stage feature fusion coarse network,the latent diffusion-driven texture refinement network and the infrared denoising module to obtain the infrared super-resolution images. In the inference phase, the real world low-quality infrared images are processed through the super-resolution network trained as described above to obtain the final infrared super-resolution images $I_{SR}$.}
\label{fig2}
\end{figure*}

Image super-resolution (SR) aims to reconstruct high-resolution (HR) images from their degraded low-resolution (LR) observations~\cite{sj1,sj2,sj3,sj4}. Many methods have been proposed to address the problem of image super-resolution. LF-DEST~\cite{xiao2025incorporating} significantly enhances the robustness and restoration quality of light field super-resolution models in unknown, complex real-world degradation scenarios by utilizing technologies including joint estimation of blur kernels and noise maps, and designing modulated selective fusion modules. EATER~\cite{xiao2024event} utilizes lightweight alignment (EAA) and fusion (EAF) units to upgrade traditional video super-resolution models into event camera-driven ones in a parameter-efficient manner. MamEVSR~\cite{xiao2025event} leverages the iMamba block with interleaved token modeling and the cMamba block for cross-modality interactions to significantly enhance event camera video super-resolution performance while preserving linear computational complexity.
Additionally, we have also conducted research on feature extraction in related work. MIBNet~\cite{guo2025deep} leverages gate mechanism-driven variational information bottlenecks and multi-view fusion strategies to significantly enhance predictive accuracy for complex disease-related micro-ribonucleic acids in heterogeneous biological networks. GCR-Net~\cite{li2023learning} leverages the exponential gated linear unit (EGLU), residual connections (EGCRU), and bidirectional gated recurrent units (Bi-GRU) to significantly enhance predictive accuracy for translation initiation sites in genomic sequences. TinyFL~\cite{hung2025tinyfl_hkd} leverages hierarchical knowledge distillation and advanced encryption standard to significantly enhance model accuracy and computational efficiency for federated learning on edge devices while safeguarding data privacy.

\subsection{Diffusion Models}
Diffusion models are generative models that generate high-quality samples by simulating a gradual denoising process of data. In recent years, diffusion models have performed exceptionally well in generative tasks. DDPMs (Denoising Diffusion Probabilistic Models)~\cite{ddpm1} are the first to propose a generative framework based on step-by-step denoising, demonstrating its formidable capabilities in image generation. DDIM (Denoising Diffusion Implicit Model)~\cite{ddim1} has enhanced sampling efficiency and generation quality on the foundation of DDPM. In addition to image generation, diffusion models are also widely applied to SR tasks. For example, SR3~\cite{sr3} uses diffusion models to address image restoration and SR issues, combining multi-scale structures to enhance detail reconstruction. DiffBIR~\cite{diffbir} applies diffusion models to blind image super-resolution, achieving image reconstruction in real-world scenarios by combining conditional diffusion and multi-stage training methods. These methods predominantly employ large-scale network architectures, which may exhibit suboptimal convergence performance when the training data is limited and complex.
\section{Methodology}
For infrared images, we design a Gaussian probability-guided adaptive quantization method for degradation operations during model input processing. This method preserves fundamental structural information while effectively reducing noise interference and smoothing pixel-level variations during degradation, thereby significantly mitigating overfitting risks under data-scarce conditions. Subsequently, we propose a novel two-stage infrared image recovery model: The first stage employs a Transformer-based architecture to capture essential structural patterns and local detail features for preliminary resolution restoration. The second stage utilizes a Diffusion-based model integrated with denoising mechanisms to perform high-quality refinement on the initial output. The final reconstructed infrared images maintain synthesized texture details while exhibiting visually sharper and more faithful results.
\subsection{Gaussian Probability Guided Quantization}

\begin{figure*}[!t]
\centering
\includegraphics[width=0.9\linewidth]{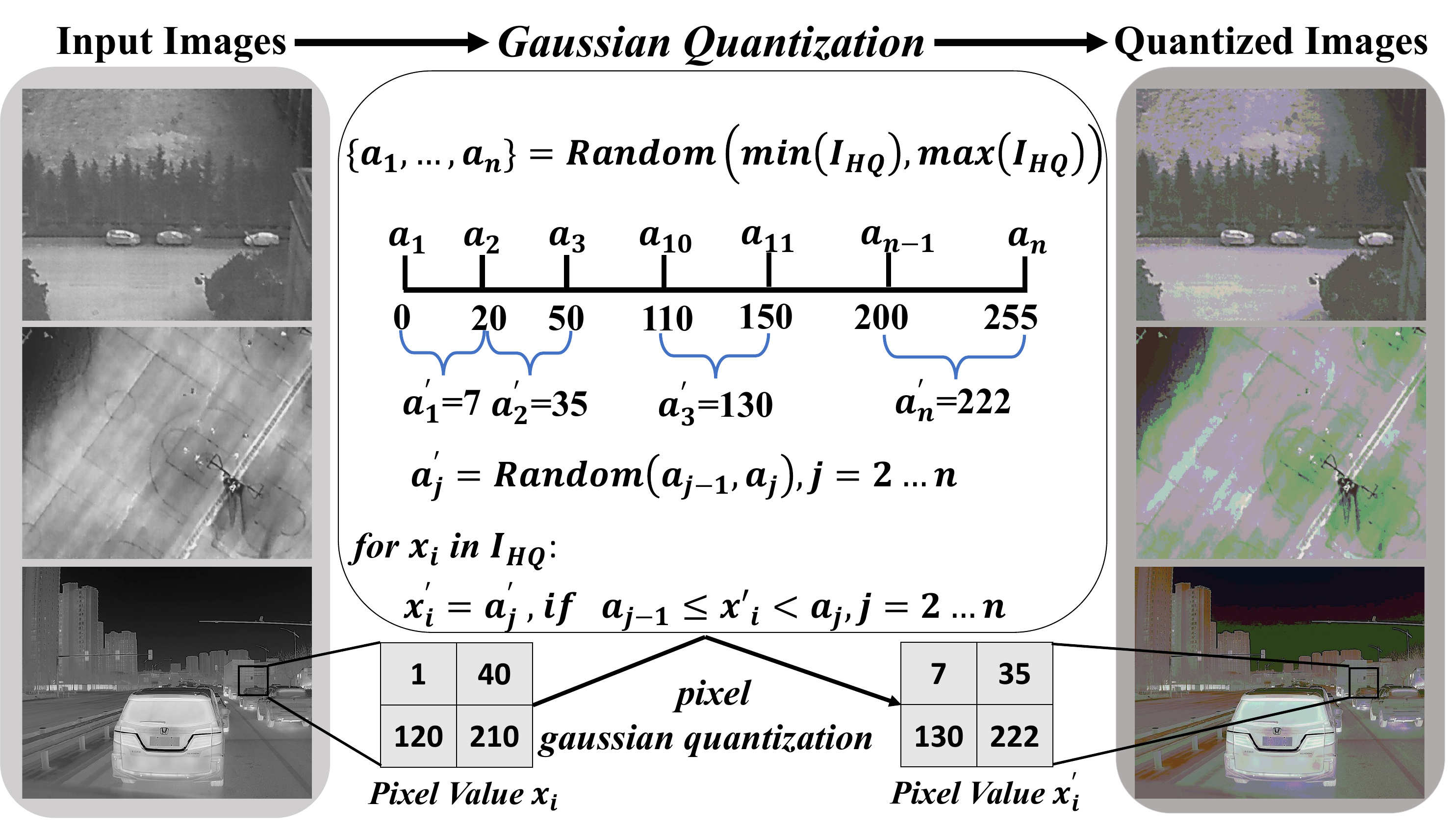}
\caption{Gaussian probability guided quantization for richer representation laerning. The pixel points in the input infrared image on the left first undergo Gaussian sampling to determine the number of quantization intervals. Afterwards, random pixel points are selected within each interval as the representative value for that interval, and all pixel values within the interval are replaced by the representative value.}
\label{fig3}
\end{figure*}

As shown in Fig.~\ref{fig2}, we propose a Gaussian probability-guided adaptive quantization method. As shown in part A of Fig.~\ref{fig2}, the Gaussian quantization uses the Gaussian distribution probability density function $p_{\text{gaussian}}$ to map and quantize the pixel values of the input infrared images $I_{\text{HR}}$. It makes the pixel value distribution in the image more uniform, thereby achieving noise smoothing and retain the main edges and structural information. 

\emph{Adaptive Region Generation.} In the sampling process, the probability density function of the Gaussian distribution $p_{\text{gaussian}}$ is defined as follows:
\begin{equation}
  p_{\text{gaussian}}(x) = \frac{1}{\sqrt{2\pi\sigma^2}} \exp\left(-\frac{(x-\mu)^2}{2\sigma^2}\right)
  \label{eq2}
\end{equation}
where $x$ represents the pixel values in the image, ranging from [0,255]. $\mu$ is the mean of the Gaussian distribution, representing the centric anchor of the distribution. The standard deviation $\sigma$ determines the width of the distribution. The resulting $p_{\text{gaussian}}(x)$ redistributes the pixel values through the weights of the probability density, thereby achieving smooth quantization. We describe the specific algorithm in Algorithm~\ref{alg1}.


\emph{Dynamic Quantization.} Intuitively, we have visualized the algorithm process in  Fig.~\ref{fig3}. For the channel of the input infrared image $I_{HQ}$, we first calculate the minimum value $min(I_{HQ})$ and the maximum value $max(I_{HQ})$ of the pixels in the channel. These two values are used to determine the range for each channel. Next, we use the aforementioned Gaussian sampling method to determine the number of intervals, and then randomly generate the boundary values for these intervals, as follows:
\begin{equation}
 \{a_1, \dots, a_n\} = \text{Random}(\min(I_{HQ}), \max(I_{HQ}))
  \label{eq3}
\end{equation}
where Random indicates a random sampling function. And $\{a_1,\dots,a_n\}$ refers to the pixel values of the region boundaries. In  Fig.~\ref{fig3}, we assume that $a_1$ = 0 ,..., $a_n$ =255. In the interval between $a_{j-1}$ and $a_j$ , we randomly generate proxy values within the region. Each pixel will be classified into a region based on its value and be assigned the proxy value of that region:
\begin{equation}
 a_j' = \text{Random}(a_{j-1}, a_j), \quad j = 2 \dots n
  \label{eq4}
\end{equation}
\begin{equation}
 \text{Random}(a_{j-1}, a_j) = a_{j-1} + (a_{j}-a_{j-1}) \times p_j \leq a_j
  \label{eq5}
\end{equation}
where $a_j'$ is the proxy value randomly generated within the interval between $a_{j-1}$ and $a_j$. In addition, $p_j$ is the probability associated with the randomly generated value. 

\emph{Value Assignment.} For example, in  Fig.~\ref{fig3}, for the first interval $a_1$ and $a_2$ , the proxy value $a_1'$ =7. Specifically, since $a_1$ =0, $a_2$ =20 and $p_1$ =0.35, by substituting into the proxy value formula, we get: $a_1' = a_{1} + (a_{2}-a_{1}) \times p_1=7$. After the above quantization process, all pixel values $x$ in the input images $I_{HQ}$ are quantized to new pixel values $x'$ , resulting in the Gaussian-quantized image $I_{\text{quant}}$. 

\begin{figure}[t]
\centering
\includegraphics[width=1\columnwidth]{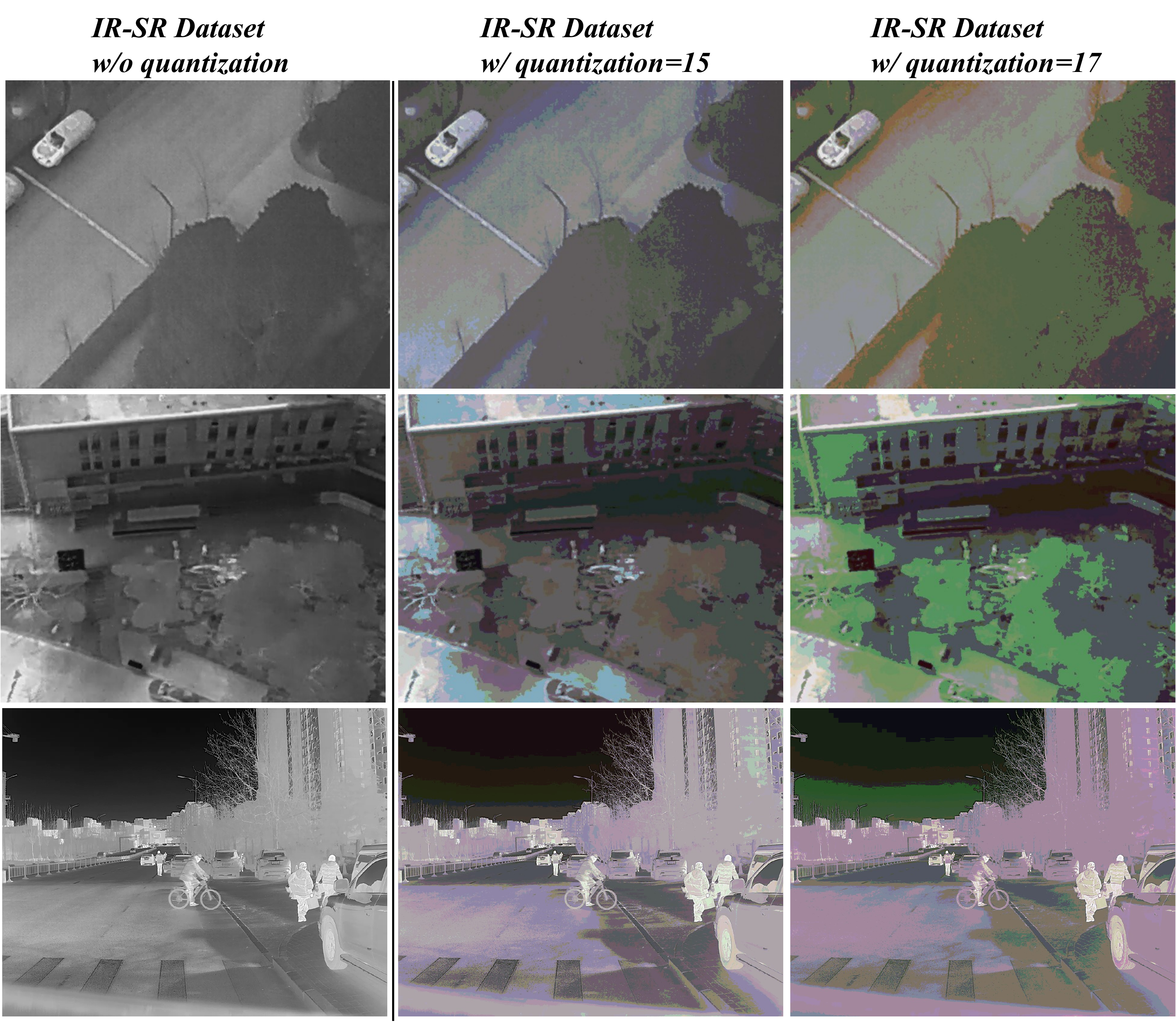}
\caption{Visualization of Gaussian quantization applied to infrared images. The left images show the original infrared image before Gaussian quantization. The middle images show the result of Gaussian quantization with 15 intervals. The right images show the result of Gaussian quantization with 17 intervals.}
\label{fig4}
\end{figure}

As shown in Fig.~\ref{fig4}, the left images are the original infrared images before Gaussian quantization. Compared to the original infrared images, the quantized images remove the high-frequency noise in the images. Our Gaussian probability-guided quantization method reduces irrelevant noise while strengthening the main structural information, helping to improve the robustness of the model. Moreover, the pixel value distribution of the quantized infrared images become smoother, which helps the network learn more stable feature representations.

\begin{algorithm}[t!]
\caption{Gaussian-guided Image Quantization}
\label{alg1}
\begin{algorithmic}

\State \textbf{Input:} Image tensor $x$, Mean $\mu=17$, Std $\sigma=2$, Range $[7, 27]$, Step size $2$
\State \textbf{Output:} Quantized image tensor $x'$

\State \textbf{Step 1: Adaptive Region Generation}
\State Generate $z \sim \mathcal{N}(\mu, \sigma^2)$ 
\State $z \leftarrow \max(7, \min(\text{round}(z), 27))$ 
\State Adjust $z$ to nearest step: $z \leftarrow z + (2 - (z-7)\bmod2)$
\State $K \leftarrow z$ 

\State \textbf{Step 2: Dynamic Quantization}
\State Compute $v_{\min}, v_{\max}$ per channel
\State Generate $\{q_1,\dots,q_{K-1}\}$ in $[v_{\min}, v_{\max}]$ 
\State Define regions: $[v_{\min}, q_1], (q_1, q_2], \dots, (q_{K-1}, v_{\max}]$

\State \textbf{Step 3: Value Assignment}
\For{each pixel $p$ in $x$}
    \State Find region $R_j$ containing $p$
    \State Quantize using strategy:
    \If{strategy == "middle"}
        \State $p' \leftarrow \text{mid}(R_j)$
    \ElsIf{strategy == "random"} 
        \State $p' \leftarrow \text{Uniform}(R_j.\text{left}, R_j.\text{right})$
    \ElsIf{strategy == "zero"}
        \State $p' \leftarrow 0$ with probability $1-p$
    \EndIf
\EndFor

\State \Return Quantized tensor $x'$ 

\end{algorithmic}
\end{algorithm}

\begin{algorithm}[t!]
\caption{DroneSR: Two-Stage Thermal Image SR}
\label{alg2}
\begin{algorithmic}

\State \textbf{Input:} High-quality thermal image $I_{HQ}$
\State \textbf{Output:} Reconstructed super-resolution image $I_{SR}$

\State \textbf{Degradation Process}
\State Apply quantization degradation
\State $I_{quant} \leftarrow \text{GaussianQuantization}(I_{HQ})$ 

\State \textbf{Stage 1: Coarse Reconstruction}
\State Multi-stage Feature Fusion Coarse Network
\State $I_{coarse} \leftarrow \text{CoarseNet}(I_{quant})$ 

\State \textbf{Stage 2: Refined Reconstruction}
\State Latent Diffusion-driven Texture Refinement Network
\State $I_{refine} \leftarrow \text{RefineNet}(I_{coarse})$ 
\State Infrared Denoising Module
\State $I_{SR} \leftarrow
\text{IDM}(I_{refine})$ 
\State \Return $I_{SR}$
\end{algorithmic}
\end{algorithm}

\subsection{Multi-stage Feature Fusion Coarse Network}
To implement thermal image super-resolution with Gaussian quantized image, we input $I_{\text{quant}}$ into the multi-stage feature fusion coarse network for training. It is an image reconstruction network built upon the Swin Transformer architecture. The reconstruction network is mainly divided into three stages: shallow feature extraction, deep feature extraction, and image reconstruction. Specifically, the input Gaussian-quantized image $I_{\text{quant}}$ first passes through a 3$\times$3 convolutional layer to extract shallow features, capturing the basic structural information and local detail features in the image. Next, the shallow features of the image are passed into a residual Swin Transformer module. This module performs nonlinear transformations and deep feature extraction to recover the high-resolution feature representation of the image.
As shown in part B of  Fig.~\ref{fig2}, after deep feature extraction, the network proceeds to the image reconstruction stage. The fused features are upsampled and processed through convolution operations to restore spatial resolution, resulting in the reconstructed super-resolution image $I_{\text{coarse}}$. The network uses the $\mathcal{L}_{\text{2}}$ loss function to measure and minimize the error between the reconstructed image $I_{\text{coarse}}$ and the input Gaussian quantized image $I_{\text{quant}}$:
\begin{equation}
 \mathcal{L}_{MFFCN} \ = \| I_{coarse} - I_{quant} \|_2^2
  \label{eq6}
\end{equation}
where $\| {\cdot } \|_2^2$ represents the squared Euclidean distance. This component completes the preliminary recovery of the infrared image.

\subsection{Latent Diffusion-driven Texture Refinement Network}
To address the issue of detail blur in the coarse reconstructed image, we introduce the latent diffusion-driven texture refinement network. The network maps inputs to the latent space via a pretrained VAE encoder~\cite{vae}, while leveraging the UNet decoder~\cite{unet} to control the fidelity of image reconstruction.
As shown in part C of  Fig.~\ref{fig2}, we introduce a generative prior based on the Latent Diffusion Model (LDM)~\cite{ldm}. It generates images that better align with the true distribution by mapping the image to a latent space, performing denoising training and resampling. LDM~\cite{ldm} maps the image to a low-dimensional latent space using a pre-trained latent space encoder. This latent space effectively compresses the input data while preserving the most important semantic and structural information. Then, the diffusion model performs generative learning in this low-dimensional latent space, significantly reducing the computational overhead during the diffusion process.

First, the input image $I_{\text{coarse}}$ is encoded into a low-dimensional latent space $z_0$, by the latent space encoder (Encoder) to compress its structural and semantic information, as shown below:
\begin{equation}
 z_0=E(I_{coarse})
  \label{eq7}
\end{equation}

Next, during the forward diffusion process, the latent space representation $z_0$ is gradually corrupted by adding noise to generate a series of intermediate states $z_t$, progressively approaching the standard normal distribution $\mathcal{N}(0,I)$ , as expressed by:
\begin{equation}
 q(z_t \mid z_0) = \mathcal{N}(z_t; \sqrt{\bar{\alpha}_t} z_0, (1 - \bar{\alpha}_t) \mathbf{I}), \quad \bar{\alpha}_t = \prod_{i=1}^t \alpha_i
  \label{eq8}
\end{equation}
where $z_t$ is the t-th state during the diffusion process, and $\bar{\alpha}_t$ is the cumulative term of the noise schedule. The final diffusion result $z_t$ approaches standard Gaussian noise through multiple iterations. Afterwards, in the reverse diffusion process, LDM learns a conditional probability distribution $p_\theta(z_{t-1} \mid z_t)$ to gradually restore the noisy state $z_t$ back to the true distribution of the latent space $z'_0$. The reverse diffusion formula is as follows:
\begin{equation}
 p_\theta(z_{t-1} \mid z_t) = \mathcal{N}(z_{t-1}; \mu_\theta(z_t, t), \Sigma_\theta(z_t, t))
  \label{eq9}
\end{equation}
where $\mu_\theta$ is the mean estimated by the network, used to predict $z_{t-1}$. $\Sigma_\theta$ is the estimated covariance, which controls the uncertainty in the generated output. $\theta$ represents the parameters of the diffusion model, which are optimized through training.  At each time step $t$,the predicted mean of the model $p_\theta(z_{t-1} \mid z_t)$ is defined as:
\begin{equation}
 \mu_\theta(z_t, t) = \frac{1}{\sqrt{\alpha_t}} \left( z_t - \frac{1 - \bar{\alpha}_t}{\sqrt{1 - \alpha_t}} \epsilon_\theta(z_t, t) \right)
  \label{eq10}
\end{equation}

To optimize $\epsilon_\theta(z_t, t)$ ,we use the loss function of LDM:
\begin{equation}
 \mathcal{L}_{\text{LDTRN}} = \mathbb{E}_{z_0, \epsilon \sim \mathcal{N}(0, I), t} \left[ \|\epsilon - \epsilon_\theta(z_t, t)\|_2^2 \right]
  \label{eq11}
\end{equation}
where $\epsilon$ is the noise randomly added during the forward diffusion,and $\epsilon_\theta(z_t, t)$ is the noise predicted by the model. This loss function measures the mean squared error between the model-predicted noise $\epsilon_\theta$ and the true noise $\epsilon$. During the iterative optimization process, the diffusion model progressively synthesizes missing textural features in infrared images. Considering the potential noise interference introduced by high-frequency detail reconstruction, we incorporate the infrared denoising module (IDM)~\cite{restormer} after the texture refinement network. As an efficient Transformer-based image denoising model, it adapts well to high-resolution image processing. Through the integration of the texture refinement network and IDM~\cite{restormer}, the infrared image not only preserves the generated textural details but also eliminates noise in high-frequency components. Our model enhances the visual precision and authenticity of the super-resolved image $I_{\text{SR}}$, making it appear more refined and realistic. The overall training procedure is shown in Algorithm~\ref{alg2}. The high-quality input image $I_{HQ}$ undergoes quantization degradation to yield a low-quality image $I_{quant}$, which is then fed into the two-stage image restoration model to obtain the high-fidelity output image $I_{SR}$.

\section{Thermal Imaging Super-Resolution Benchmark}
As shown in  Fig.~\ref{figdataset}, our work introduces a real-world infrared dataset construction method, designed to address diversity gaps in thermal imaging resources and optimize for drone-based super-resolution (SR) challenges. We collect diverse infrared data from the internet, including drone-captured aerial infrared datasets and outdoor infrared datasets. In addition, we crawl ground infrared data from the IRay Infrared Image Platform. It provides 1280×1024 high-resolution infrared images, featuring dynamic urban scenes. For the training set, we randomly select 200 images, including 150 aerial images and 50 ground images. For the validation set, we select 30 aerial images and 20 ground images. For the test set, we randomly select a total of 50 images for metric evaluation. We propose the hybrid dataset that uniquely combines multi-platform acquisition (aerial and ground sensors) and multi-scale resolution images, emphasizing the diversity and limitations of thermal imaging resources.

\section{Experiment}
\subsection{Experimental Settings}
In the experiments, we focus on the challenging $\times$4 scale infrared super-resolution task, while the proposed method can also be applied to other scaling factors, such as 3 and 6. For SR models~\cite{realesrgan,stablesr} applicable to different scaling factors, we consistently use a  $\times$4 trained model for experiments at different scaling factors. Our method, along with other comparative methods, is experimentally studied on an NVIDIA RTX 3090 GPU. Our method trains for 100000 steps on the coarse reconstruction network and for 5000 steps on the texture refinement network. And we train for 100000 steps on the infrared denoising module. The batch size is set to 4. During the Gaussian probability-guided quantization phase, we set the mean $\mu$ to 17 and the standard  deviation $\sigma$ to 2. This parameterization enables diverse degradation operations during quantization iterations, generating images at multiple degradation scales. Thereby, it enhances input diversity under data-scarce training conditions, consequently improving model robustness. For iteration count determination, we observed the training processes of both stages: iteration parameters were finalized when the loss curves showed stable convergence. Regarding diffusion steps in the diffusion model, we set this value to 50 since it achieves an optimal balance between information retention and computational load.

\begin{table}[t]\rmfamily
  \centering
  \setlength{\tabcolsep}{5pt}
  \caption{Quantitative results of $\times$3 scale super-resolution on the test set compared with other methods. The best and second-best results are highlighted in red and blue.}
  \begin{tabular}{cccccc}
  \toprule
   Method & {PSNR$\uparrow$} &  {SSIM$\uparrow$} & {LPIPS$\downarrow$}  & {FSIM$\uparrow$} &  {MS-SSIM$\uparrow$} \\
    \midrule
   Real-ESRGAN~\cite{realesrgan} & 28.68 & 0.8320 & \textcolor{red}{0.2134} & 0.9171 & 0.9261 \\
   LIRSN~\cite{lirsn}                  & \textcolor{blue}{29.22} & \textcolor{blue}{0.8491} & 0.4522 & \textcolor{blue}{0.9254} & \textcolor{blue}{0.9439}\\
   DiffBIR~\cite{diffbir}                  & 26.56 & 0.7875 & 0.3332 & 0.8548 & 0.8675 \\
    StableSR~\cite{stablesr}              & 25.62 & 0.7329 & \textcolor{blue}{0.2914} & 0.8931 & 0.9114\\
     \hline \hline
    \textbf{Ours}                     & \textcolor{red}{29.88}        & \textcolor{red}{0.8697}        & 0.3376            & \textcolor{red}{0.9322}     & \textcolor{red}{0.9502}        \\
    \bottomrule
  \end{tabular}
  \label{table2}
\end{table}

\begin{table}[t]\rmfamily
  \centering
  \setlength{\tabcolsep}{5pt}
  \caption{Quantitative results of $\times$4 scale super-resolution on the test set compared with other methods. The best and second-best results are highlighted in red and blue.}
  \begin{tabular}{cccccc}
  \toprule
   Method & {PSNR$\uparrow$} &  {SSIM$\uparrow$} & {LPIPS$\downarrow$}  & {FSIM$\uparrow$} &  {MS-SSIM$\uparrow$} \\
    \midrule
   Real-ESRGAN~\cite{realesrgan}            & 27.71 & 0.8097 & \textcolor{red}{0.2399}  &
    0.8996 & 0.9040\\
    LIRSN~\cite{lirsn}                  & 27.47 & 0.8067 & 0.5481 & 0.8857 & 0.9069\\
    DiffBIR~\cite{diffbir}                  & 27.08 & 0.7882 & \textcolor{blue}{0.2812} &  0.8713 & 0.8773 \\
    StableSR~\cite{stablesr}              & 25.41 & 0.7115 & 0.3200 &  0.8797 & 0.8973\\
    Resshift~\cite{resshift}            & \textcolor{blue}{28.37} & \textcolor{blue}{0.8183} & 0.3488  &
    \textcolor{blue}{0.9119} & \textcolor{blue}{0.9256}\\
    \hline \hline
    \textbf{Ours}                     & \textcolor{red}{30.22}        & \textcolor{red}{0.8618}        & 0.3622     & \textcolor{red}{0.9303}     & \textcolor{red}{0.9445}        \\
    \bottomrule
  \end{tabular}
  \label{table1}
\end{table}

\begin{table}[t]\rmfamily
  \centering
  \setlength{\tabcolsep}{5pt}
  \caption{Quantitative results of $\times$6 scale super-resolution on the test set compared with other methods. The best and second-best results are highlighted in red and blue.}
  \begin{tabular}{cccccc}
  \toprule
   Method & {PSNR$\uparrow$} &  {SSIM$\uparrow$} & {LPIPS$\downarrow$}  & {FSIM$\uparrow$} &  {MS-SSIM$\uparrow$} \\
   \midrule
   Real-ESRGAN~\cite{realesrgan} & 25.65 & 0.7673 & \textcolor{blue}{0.3269} & 0.8477 & 0.8391 \\
   LIRSN~\cite{lirsn}                  & 25.89 & 0.7735 & 0.6030 & 0.8317 & 0.8567\\
   DiffBIR~\cite{diffbir}                  & \textcolor{blue}{26.42} & \textcolor{blue}{0.7771} & \textcolor{red}{0.3102} &  \textcolor{blue}{0.8539} & \textcolor{blue}{0.8595} \\
    StableSR~\cite{stablesr}              & 24.91 & 0.6745 & 0.4194 &  {0.8358} & 0.8462\\
     \hline \hline
    \textbf{Ours}                     & \textcolor{red}{27.41}        & \textcolor{red}{0.8052}        & 0.4828        & \textcolor{red}{0.8816}     & \textcolor{red}{0.8980}        \\
    \bottomrule
  \end{tabular}
  \label{table3}
\end{table}

\subsection{Comparison with Baseline Methods}
To validate the effectiveness of our method, we compare it with several state-of-the-art methods, including Real-ESRGAN~\cite{realesrgan}, LIRSN~\cite{lirsn}, DiffBIR~\cite{diffbir}, StableSR~\cite{stablesr} and Resshift~\cite{resshift}. For Real-ESRGAN~\cite{realesrgan} , LIRSN~\cite{lirsn} and DiffBIR~\cite{diffbir} , we obtain the actual quantitative results by training and reproducing them using the official code. For StableSR~\cite{stablesr} and Resshift~\cite{resshift} , we test them using the official code and provided models. The comparative quantitative results with other methods are summarized in Table~\ref{table2}, Table~\ref{table1} and Table~\ref{table3}. Since Resshift~\cite{resshift} is effective for $\times$4 scale super-resolution, we do not compare its performance at other resolution scales.

We first perform a quantitative comparison on the $\times$4 scale super-resolution test set. As shown in Table~\ref{table1}, our method outperforms existing super-resolution methods on several perceptual metrics, including PSNR, SSIM, FSIM and MS-SSIM. Specifically, on the PSNR metric, our Gaussian quantization-based method achieves the highest score of 30.22. Our method outperforms Resshift~\cite{resshift} by 1.85 dB in PSNR, and the SSIM value increases by 0.0435. Although Real-ESRGAN~\cite{realesrgan} achieve the best result in LPIPS, their performance in other evaluation metrics is still significantly lower than ours. This highlights the advantages of our Gaussian probability-guided quantization method.

\begin{figure*}[!t]
\centering
\includegraphics[width=\linewidth]{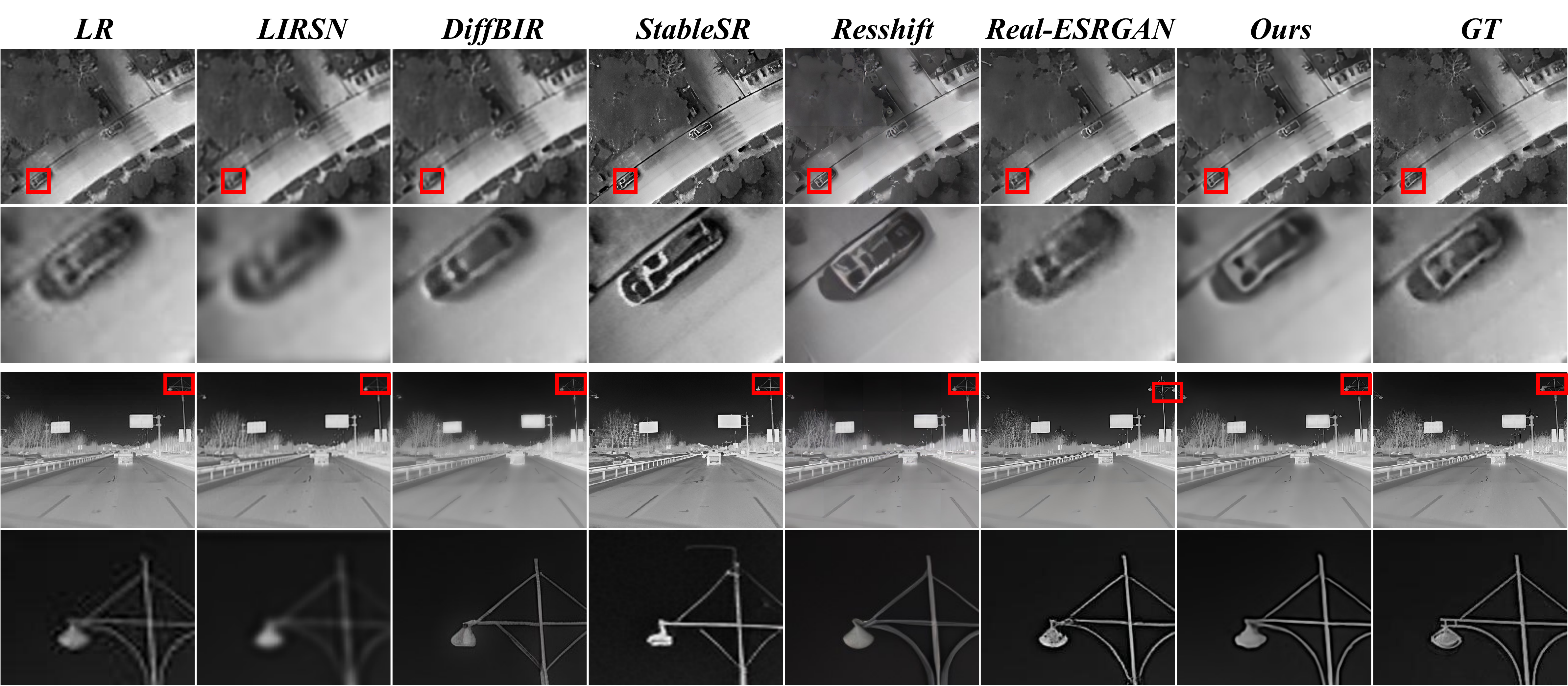}
\captionsetup{justification=centering}
\caption{Visualization results of $\times$4 scale super-resolution.}
\label{fig5}
\end{figure*}

\begin{figure}[!t]
\centering
\includegraphics[width=\linewidth]{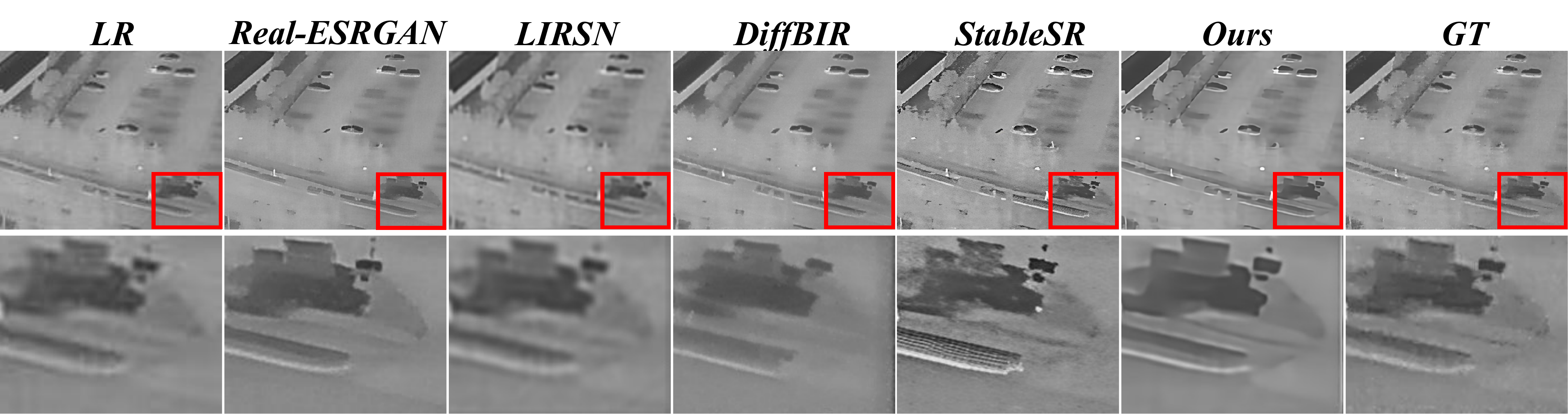}
\captionsetup{justification=centering}
\caption{Visualization results of $\times$3 scale super-resolution. }
\label{fig6}
\end{figure}

\begin{figure}[!t]
\centering
\includegraphics[width=\linewidth]{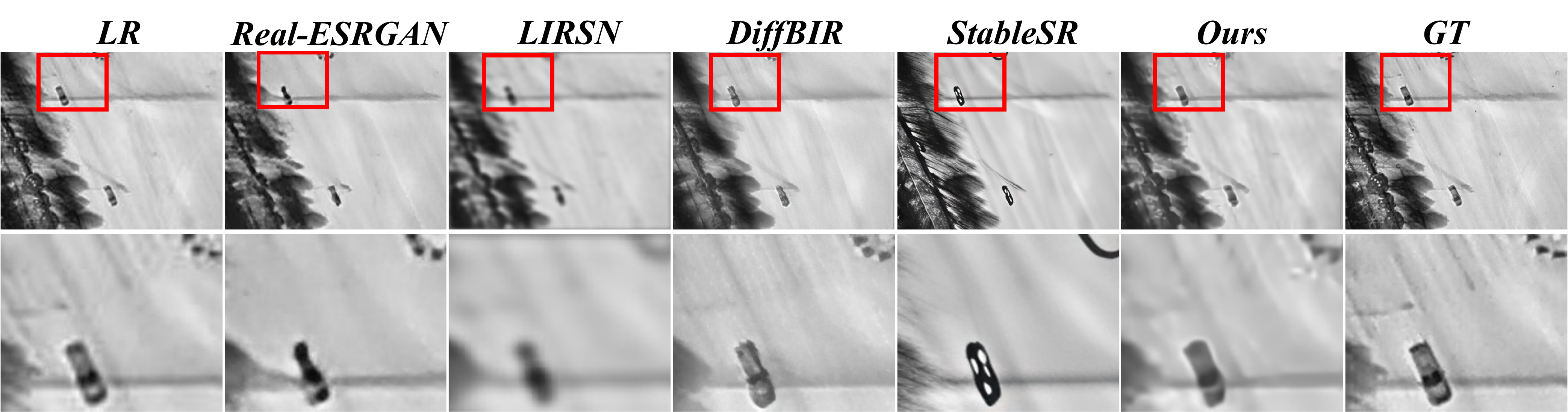}
\captionsetup{justification=centering}
\caption{Visualization results of $\times$6 scale super-resolution. }
\label{fig7}
\end{figure}

\begin{figure}[!t]
\centering
\includegraphics[width=\linewidth]{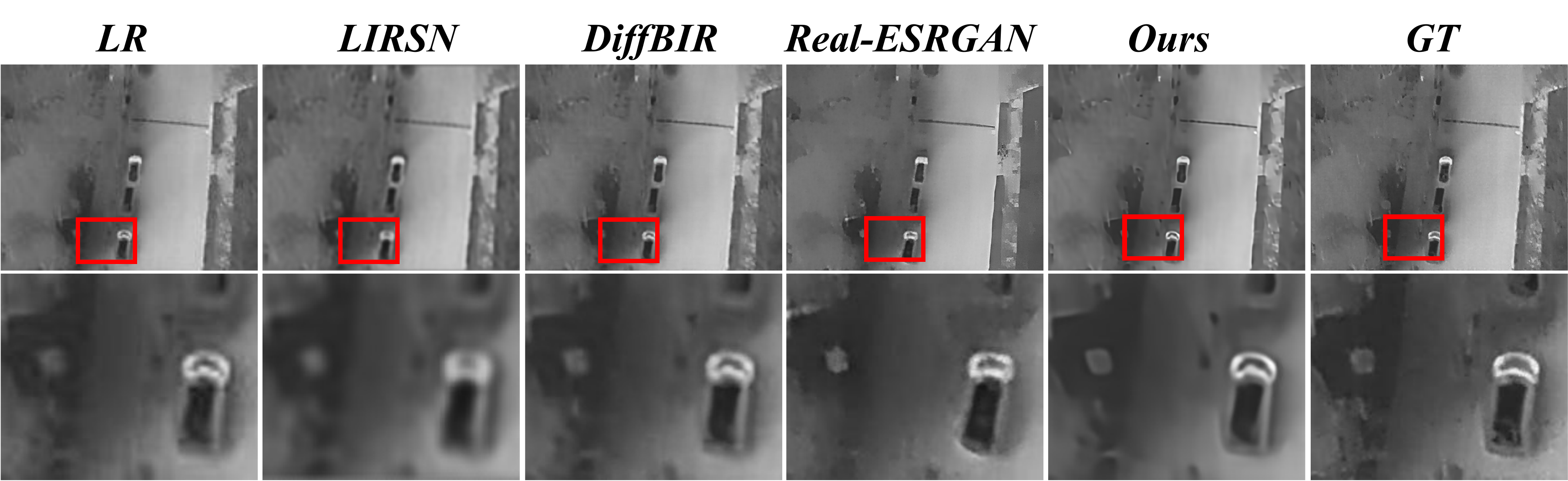}
\captionsetup{justification=centering}
\caption{Underperforming visualization results of $\times$4 scale super-resolution.} 
\label{fig8}
\end{figure}

To demonstrate the effectiveness of our method, we present the visual results of the test set with a $\times$4 scale super-resolution in  Fig.~\ref{fig5}. The first column shows the LR image obtained by downsampling the original image by a factor of four. The second, third, fourth, fifth and sixth columns correspond to the images obtained using LIRSN~\cite{lirsn}, DiffBIR~\cite{diffbir}, StableSR~\cite{stablesr}, Resshift~\cite{resshift} and Real-ESRGAN~\cite{realesrgan}, respectively. The seventh column shows the infrared super-resolved image obtained by our method, and the last column shows the real infrared image. As shown in  Fig.~\ref{fig5}, our method significantly improves the visual quality of the images. Specifically, the sixth column of  Fig.~\ref{fig5} and its enlarged local image demonstrate that our method can generate realistic details.

In addition to the $\times$4 scale super-resolution experiments, we also conduct corresponding studies on $\times$3 and $\times$6 scale super-resolution to demonstrate the effectiveness of our method. As shown in Table~\ref{table2} and Table~\ref{table3}, our method also outperforms existing super-resolution methods across multiple perceptual metrics. In Table~\ref{table2}, our method achieves the best evaluation results across four metrics, including PSNR, SSIM, FSIM and MS-SSIM. In the quantitative results of Table~\ref{table2}, compared to DiffBIR~\cite{diffbir}, our method improves by 3.32 dB in the PSNR metric and about 0.08 in the SSIM metric. Compared to other methods, our quantitative metrics also show significant improvement.

We present the visual results of the test set at $\times$3 scale super-resolution in Fig.~\ref{fig6}. From Fig.~\ref{fig6}, it can be seen that the visual quality of the images is also significantly improved after applying our method. Specifically, as shown in the sixth column of Fig.~\ref{fig6} and its enlarged local image, our method is able to generate realistic details. The infrared images produced by LIRSN~\cite{lirsn} exhibit blurring effects, while the infrared images generated by DiffBIR~\cite{diffbir} contain some artifacts. We present the visualization results of the test set with $\times$6 scale super-resolution in Fig.~\ref{fig7}. As shown in the sixth column of Fig.~\ref{fig7}, our method is capable of generating smoother images.

As shown in Fig.~\ref{fig8}, all methods demonstrate unsatisfactory results when performing super-resolution tasks on low-quality images where information loss occurs due to fast-moving targets. We zoom in on the distorted area in the image, which originates from motion blur of moving objects causing the loss of original information. We will conduct further research to improve related methodologies in future work.

\textbf{Efficiency analysis}. In terms of parameter count, we categorize them by different modules. The parameter count of the coarse network is 15.8 M, the diffusion model is 865.9 M, and the infrared denoising module is 26.1 M. As shown in Table~\ref{table4}, we examine the model memory usage and inference time per image for different methods compared to our method during the inference stage. Although LIRSN~\cite{lirsn} requires the least memory and has the shortest inference time, its simple model structure leads to suboptimal final quantitative results. Our method requires the most memory and inference time, but due to its model stability and the larger sampling steps set during the inference phase, we achieve the best results in both quantitative metrics and qualitative outcomes.

\begin{table}[h]\rmfamily
  \centering
  \setlength{\tabcolsep}{2.5pt}
  \small
  \caption{Efficiency analysis on $\times$4 scale super-resolution test set.}
  \begin{tabular}{@{}c c c c c c@{}}
    \toprule
   Method & Memory& Time & GFLOPS&PSNR&SSIM \\
    \midrule
  LIRSN~\cite{lirsn} & 2 GB & 0.1 s & 0.6&27.47&0.8067  \\ Real-ESRGAN~\cite{realesrgan} & 4 GB & 0.5 s &90&27,71&0.8097  \\ 
            StableSR~\cite{stablesr} & 9.2 GB & 1.13 s & 100&25.41&0.7115  \\
           DiffBIR~\cite{diffbir} & 16 GB & 1.1 s & 220 & 27.08 & 0.7882  \\
             \hline \hline
            \textbf{Ours} & 16 GB & 1.2 s & 300 & 30.22&0.8618 \\
    \bottomrule
  \end{tabular}
  \label{table4}
\end{table}

\subsection{Ablation Study}
To demonstrate the effectiveness of the proposed method, we conduct the ablation study. This study analyzes the quantitative results of the Gaussian quantization method, different resampling steps during the inference phase and the role of the denoiser.

\begin{table}[ht]\rmfamily
  \centering
  \setlength{\tabcolsep}{2pt}
  \small
  \caption{Quantitative results of using and not using the Gaussian quantization method. }
  \begin{tabular}{ccc}
    \toprule
    Metrics & w/o Gaussian Quantization &   w/ Gaussian Quantization \\
    \midrule
    {PSNR$\uparrow$} & 26.80 & \textbf{29.57} \\
    {SSIM$\uparrow$} & 0.7781 & \textbf{0.8390}\\
    {LPIPS$\downarrow$} &\textbf{0.2805} & 0.3270\\
    {FSIM$\uparrow$} & 0.8630 & \textbf{0.9288}\\
    {MS-SSIM$\uparrow$}& 0.8648 & \textbf{0.9350}\\
    \bottomrule
  \end{tabular}
  \label{table5}
\end{table}

\begin{table}[h]\rmfamily
  \centering
  \setlength{\tabcolsep}{5pt}
  \normalsize
  \caption{Quantitative results of different numbers of steps in the inference phase of the diffusion model. We perform this experiment on $\times$4 scale super-resolution test set. The best and second-best results are highlighted in red and blue.}
  \begin{tabular}{cccccc}
  \toprule
   Steps & {PSNR$\uparrow$} &  {SSIM$\uparrow$} & {LPIPS$\downarrow$}  & {FSIM$\uparrow$} &  {MS-SSIM$\uparrow$} \\
   \midrule
   10 & \textcolor{red}{29.65} & \textcolor{red}{0.8420} & 0.3418  & \textcolor{blue}{0.9286} & \textcolor{red}{0.9366} \\
    20 & \textcolor{blue}{29.62} & \textcolor{blue}{0.8408} & 0.3366 & 0.9288 & \textcolor{blue}{0.9360}\\
    30 & 29.56 & 0.8395 & 0.3316  & 0.9285 & 0.9352\\
    40 & 29.56 & 0.8388 & \textcolor{red}{0.3265}  & 0.9288 & 0.9349\\
    50 & 29.57 & 0.8390 & \textcolor{blue}{0.3270} &   \textcolor{red}{0.9288} & 0.9350\\
    \bottomrule
  \end{tabular}
  \label{table6}
\end{table}

\begin{table}[ht]\rmfamily
  \centering
  \setlength{\tabcolsep}{5pt}
  \caption{Quantitative results of using and not using the infrared denoising module. We perform this experiment on $\times$4 scale super-resolution test set.}
  \small
  \begin{tabular}{ccc}
    \toprule
    Metrics & w/o Infrared Denoising &  w/ Infrared Denoising\\
    \midrule
    {PSNR$\uparrow$} & 29.57 & \textbf{30.22} \\
    {SSIM$\uparrow$} & 0.8390 & \textbf{0.8618}\\
    {LPIPS$\downarrow$} &\textbf{0.3270} & 0.3622\\
    {FSIM$\uparrow$} & 0.9288 & \textbf{0.9303}\\
    {MS-SSIM$\uparrow$}& 0.9350 & \textbf{0.9445}\\
    \bottomrule
  \end{tabular}
  \label{table7}
\end{table}

Table~\ref{table5} shows the quantitative results with and without using the Gaussian quantization method. It can be seen that after using the Gaussian probability-guided quantization method, our method improves PSNR by 2.77 dB, SSIM by approximately 0.06, and shows more significant improvements in FSIM and MS-SSIM. The LPIPS metric do not show significant differences compared to the results without the Gaussian quantization method. Therefore, the Gaussian quantization method we proposed significantly enhances the robustness of the model in the infrared image super-resolution task.

In the inference stage, the number of time steps in the latent diffusion-driven texture refinement network also affects the results on the test set. As shown in Table~\ref{table6}, we analyze the quantitative results with different step counts of 10, 20, 30, 40, and 50 in the inference phase. It can be observed that when the step count is set to 10, the overall quantitative results are optimal. This also aligns with the stability of noise sampling in diffusion models.

We further conduct an ablation study on the infrared denoising module (IDM). Table~\ref{table7} demonstrates the quantitative comparisons between configurations with and without the infrared denoising module. The implementation of the IDM shows significant metric improvements, achieving a 0.65dB PSNR gain and 0.023 SSIM increase compared to configurations without the IDM. And LPIPS maintains comparable performance across both configurations. These results confirm that the IDM effectively enhances model robustness in infrared image super-resolution tasks.

\section{Conclusion}
Under the condition of limited and diverse data, infrared images are prone to overfitting during the training process of large-scale models. To address this issue, we propose a Gaussian probability-guided quantization method for image reconstruction tasks based on diffusion models. The method effectively suppresses the overfitting issue of few-shot infrared images in large-scale model training. In addition, we have built a novel multi-source infrared image dataset to support model training and evaluation. Experimental results show that our method outperforms existing diffusion models in evaluation metrics and exhibits realistic, detailed results in visualizations.

\bibliographystyle{IEEEtran}
\bibliography{ref}

\begin{IEEEbiography}
[{\includegraphics[width=1in,height=1.25in,clip,keepaspectratio]{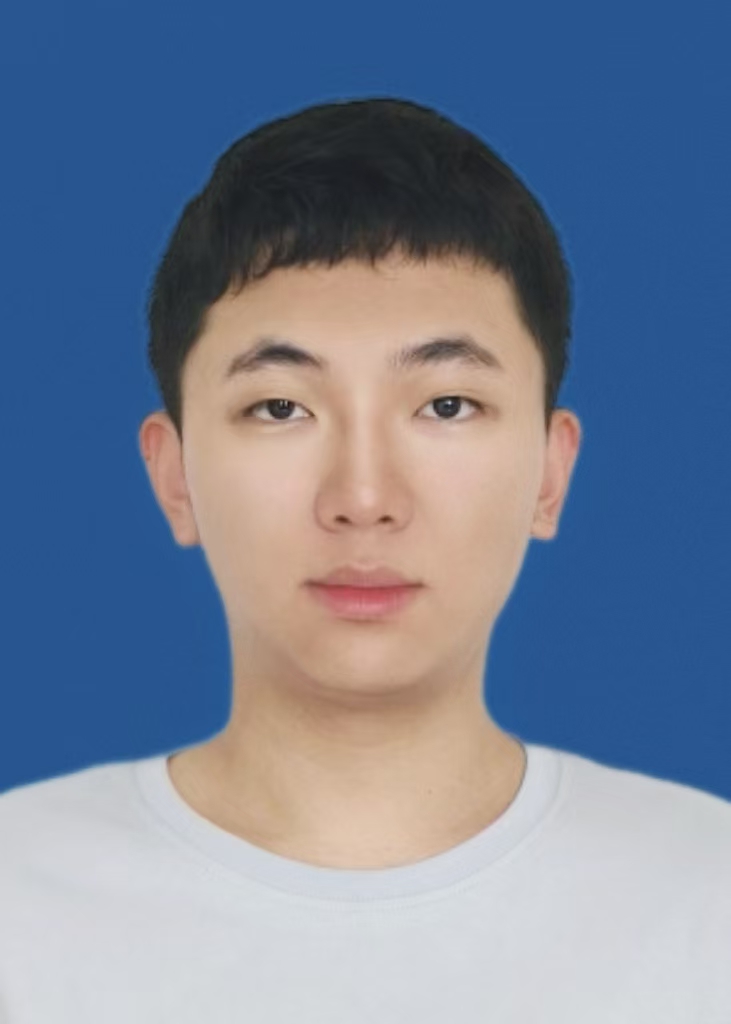}}]{Zhipeng Weng } received the B.S. degree in 2023, from the School of Information Engineering, Guangdong University of Technology, Guangzhou, China, where he is currently working towards a M.S. degree. His research interests include computer vision and machine learning
\end{IEEEbiography}

\begin{IEEEbiography}
[{\includegraphics[width=1in,height=1.25in,clip,keepaspectratio]{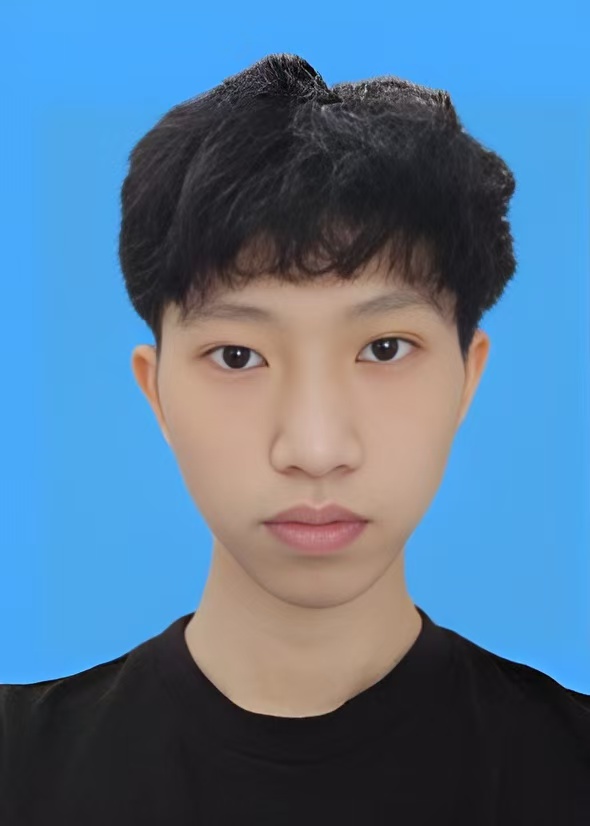}}]{Xiaopeng Liu } received the B.S. degree in 2024, from the School of Information Engineering, Guangdong University of Technology, Guangzhou, China, where he is currently working towards a M.S. degree. His research interests include computer vision and machine learning
\end{IEEEbiography}

\begin{IEEEbiography}
[{\includegraphics[width=1in,height=1.25in,clip,keepaspectratio]{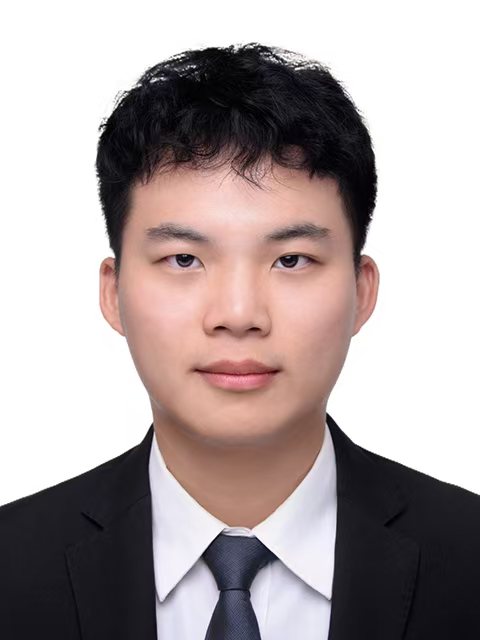}}]{Ce Liu } received the B.S. degree in 2025, from the School of Information Engineering, Guangdong University of Technology, Guangzhou, China, where he is currently working towards a M.S. degree. His research interests include computer vision and machine learning
\end{IEEEbiography}

\begin{IEEEbiography}[{\includegraphics[width=1in,height=1.25in,clip,keepaspectratio]{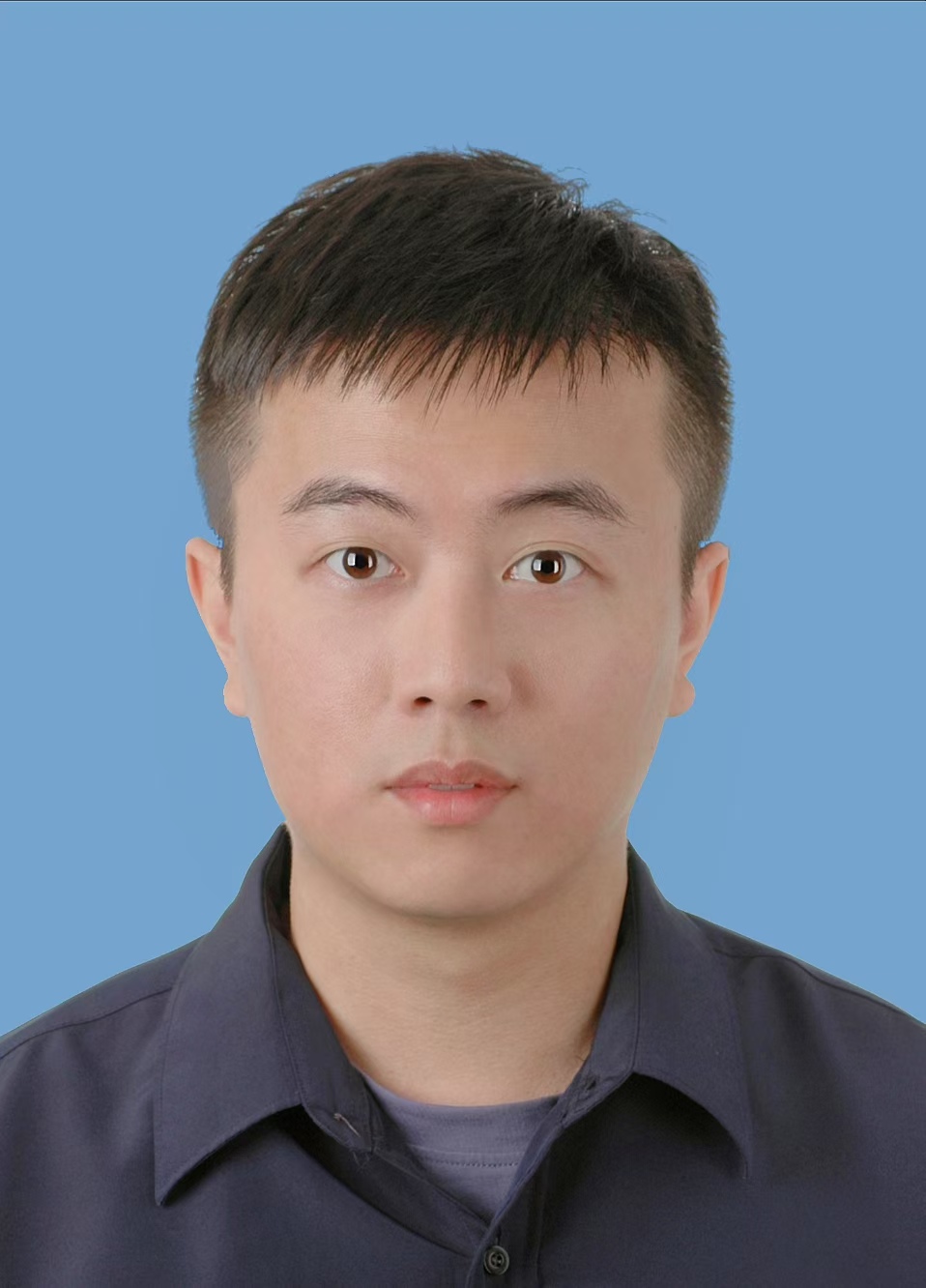}}]{Xingyuan Guo} received the B.S. degree in 2010, from the School of Automation, Guangdong University of Technology, Guangzhou, China. He is currently a senior engineer with the Southern Power Grid Company, Ltd., China. His current research interests include modeling, stability and control of industrial systems.
\end{IEEEbiography}

\begin{IEEEbiography}
[{\includegraphics[width=1in,height=1.25in,clip,keepaspectratio]{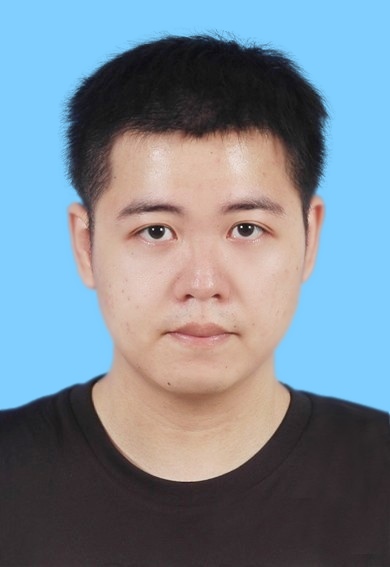}}]{Yukai Shi}
received the Ph.D. degrees from the school of Data and Computer Science, Sun Yat-sen University, Guangzhou China, in 2019. He is currently an associate professor at the School of Information Engineering, Guangdong University of Technology, China. His research interests include computer vision and machine learning.
\end{IEEEbiography}

\begin{IEEEbiography}
[{\includegraphics[width=1in,height=1.25in,clip,keepaspectratio]{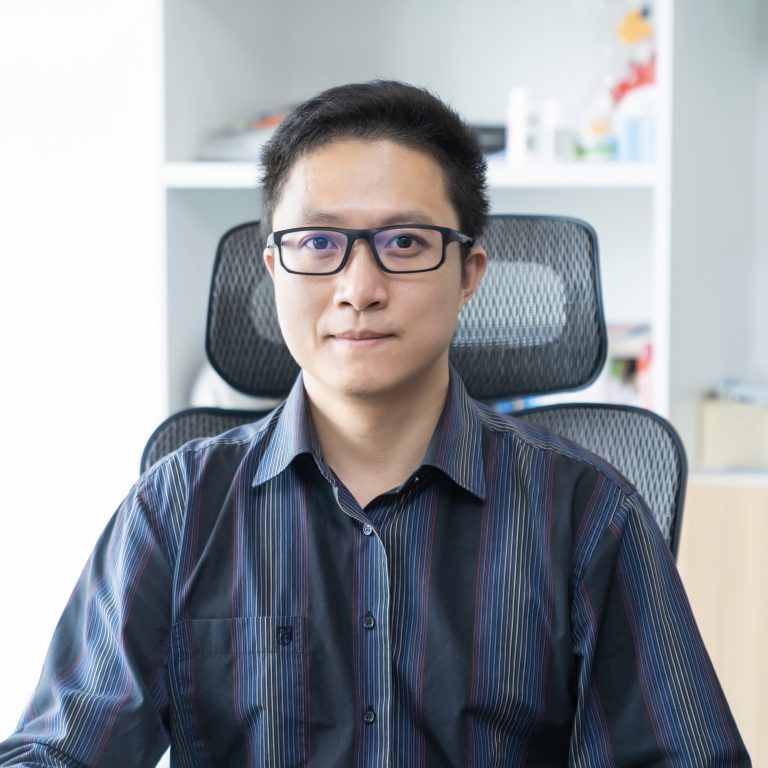}}]{Liang Lin}(Fellow, IEEE) is a Full Professor of computer science at Sun Yat-sen University. He is an associate editor of IEEE T-NNLS and IEEE T-MM, and served as Area Chairs for numerous conferences such as CVPR, ICCV, SIGKDD and AAAI. He is the recipient of numerous awards and honors including Wu Wen-Jun Artificial Intelligence Award, ICCV Best Paper Nomination in 2019, Annual Best Paper Award by Pattern Recognition (Elsevier) in 2018, Google Faculty Award in 2012.
\end{IEEEbiography}

\end{document}